\documentclass[10pt,twocolumn,letterpaper]{article}

\usepackage{cvpr}
\usepackage{times}
\usepackage{epsfig}
\usepackage{graphicx}
\usepackage{amsmath}
\usepackage{amssymb}
\usepackage{subfig}
\usepackage{graphicx}
\usepackage{balance}
\usepackage{multirow}

\usepackage[breaklinks=true,bookmarks=false]{hyperref}

\cvprfinalcopy % *** Uncomment this line for the final submission

\def\cvprPaperID{**} % *** Enter the CVPR Paper ID here

\begin{document}

%%%%%%%%% TITLE
\title{Incremental Classifier Learning with Generative Adversarial Networks}

\author{Yue Wu$^1$\quad Yinpeng Chen$^{2}$\quad Lijuan  Wang$^{2}$\quad Yuancheng  Ye$^{3}$\\ Zicheng  Liu$^{2}$\quad Yandong  Guo$^{2}$\quad Zhengyou  Zhang$^{2}$\quad Yun Fu$^{1}$\\
	$^1$Northeastern University\quad $^2$Microsoft Research\quad  $^3$City University of New York\\
	{\tt\small \{yuewu,yunfu\}@ece.neu.edu, yye@gradcenter.cuny.edu}\\ {\tt\small\{yiche,lijuanw,zliu,yandong.guo,zhang\}@microsoft.com }
	%\and
}
%yandong.guo@microsoft.com
%	Institution1\\
%	Institution1 address\\
%	{\tt\small firstauthor@i1.org}
%	% For a paper whose authors are all at the same institution,
%	% omit the following lines up until the closing ``}''.
%	% Additional authors and addresses can be added with ``\and'',
%	% just like the second author.
%	% To save space, use either the email address or home page, not both
%	\and
%	Second Author\\
%	Institution2\\
%	First line of institution2 address\\
%	{\tt\small secondauthor@i2.org}
%}

\maketitle
%\thispagestyle{empty}

%%%%%%%%% ABSTRACT
\begin{abstract}
	In this paper, we address the incremental classifier learning problem, which suffers from catastrophic forgetting. The main reason for catastrophic forgetting is that the past data are not available during learning. Typical approaches keep some exemplars for the past classes and use distillation regularization to retain the classification capability on the past classes and balance the past and new classes.  However, there are four main problems with these approaches. First, the loss function is not efficient for classification. Second, there is unbalance problem between the past and new classes. Third, the size of pre-decided exemplars is usually limited and they might not be distinguishable from unseen new classes.   Forth, the exemplars may not be allowed to be kept for a long time due to privacy regulations. To address these problems, we propose (a) a new loss function to combine the cross-entropy loss and distillation loss, (b) a simple way to estimate and remove the unbalance between the old and new classes , and (c) using Generative Adversarial Networks (GANs) to generate historical data and select representative exemplars during generation. We believe that the data generated by GANs have much less privacy issues than real images because GANs do not directly copy any real image patches. 
	We evaluate the proposed method on CIFAR-100, Flower-102, and MS-Celeb-1M-Base datasets and extensive experiments demonstrate the effectiveness of our method.
\end{abstract}

%\newpage
%%%%%%%%% BODY TEXT
\section{Introduction}
So far, most of the large-scale machine learning systems are trained in batch mode where the data samples from all the classes must be available during training. As pointed out by~\cite{rebuffi2016icarl,li2016learning}, natural learning systems are inherently incremental where new knowledge is continuously learned over time while existing knowledge is maintained. Furthermore, many computer vision applications in the real world require incremental learning capabilities because the batch-mode training does not scale up well when new classes need to be learned frequently over time.

One main problem with the incremental learning is the catastrophic forgetting~\cite{McCloskey-Cohen-PLM-1989} where the classification accuracy for the existing classes may quickly deteriorate as the new classes are added. Catastrophic forgetting is mainly caused by the fact that the past data are not available during training. 
The key challenge is how to integrate the old classifier and new data to learn the representation efficiently. Existing solutions \cite{li2016learning, rebuffi2016icarl} either use the old classifier as regularization, or select a few exemplars to represent old training data. However, neither of them can retain the classification performance on the old classes and at the same time balance the old/new classes. In addition, selecting exemplars has other limitations (e.g. not scalable, privacy issue, etc).  

We propose a new loss function to integrate cross-entropy loss and distillation loss on both old exemplars and new training samples. This allows accurate classification within old classes or new classes. To balance between old and new classes, we found that a scalar can efficiently represent the bias on new data. Thus we can simply estimate the bias on validation set and remove it from the model. The experiment results show the bias estimation is very stable across validation and test datasets.  This method outperforms the-state-of-arts iCaRL \cite{rebuffi2016icarl} on CIFAR-100 \cite{krizhevsky2009learning}.

We also investigate using GANs\cite{goodfellow2014generative,ACGAN2017,CGAN-2014,RadfordMC15,ArjovskyCB17} to replace the exemplars as it is more scalable and has less privacy issue than keeping the past data. Although GANs have limitations on image quality and mode dropping, our method outperforms the-state-of-arts LwF \cite{li2016learning} which does not use any exemplars on CIFAR-100 \cite{krizhevsky2009learning}, Flower-102 \cite{nilsback2008automated} and MS-Celeb-1M-Base \cite{lowshotface,guo2016msceleb} datasets.

\begin{figure*}[h]
	\begin{center}
		\includegraphics[width=\linewidth]{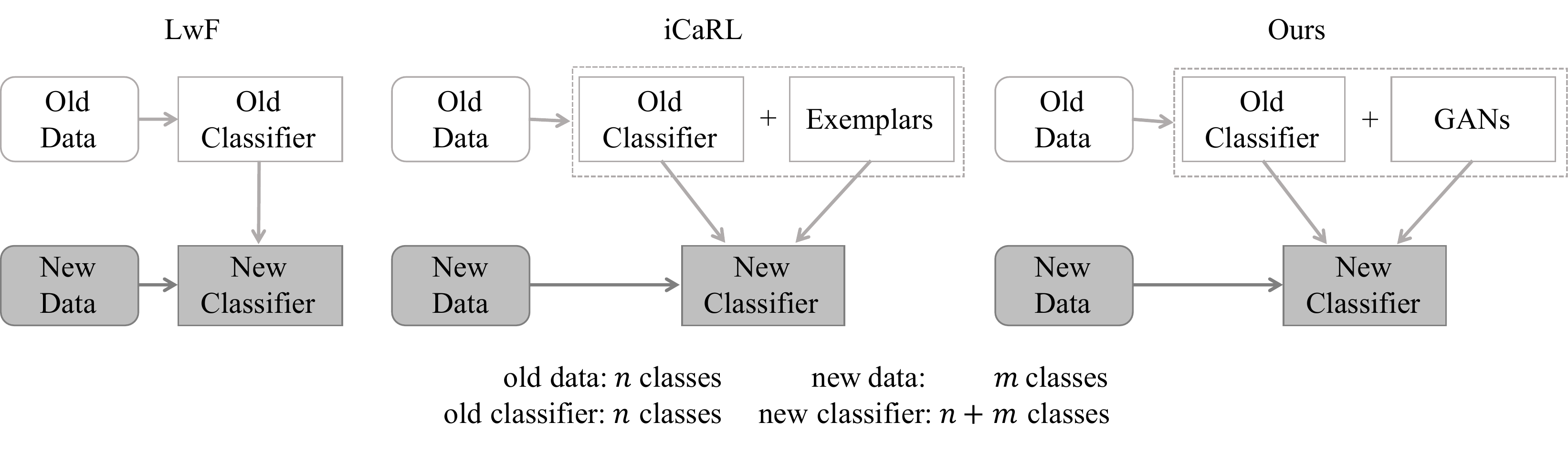}
	\end{center}
	%	\vspace{-4mm}
	\caption{Comparison of our framework with LwF \cite{li2016learning} and iCaRL \cite{rebuffi2016icarl}. LwF applies the distilling loss to the new data and iCaRL keeps a small set of exemplars from old data. Our method utilizes GANs to generated old data. }
	\label{fig:compareFramework}
	%	\vspace{-4mm}
\end{figure*}

\section{Related Work}

Incremental learning has been a long standing problem in machine learning~\cite{cauwenberghs2001incremental,polikar2001learn++,mensink2013distance,kuzborskij2013n}.  Before the deep learning took off, people had been developing incremental learning techniques by leveraging linear classifiers, ensemble of weak classifiers, nearest neighbor classifiers, etc. Recently, thanks to the exciting progress in deep learning, there has been a lot of research on incremental learning with deep neural network models. The work can be roughly divided into two categories depending on whether they require the old data or not. 

The methods in the first category do not require any old data. \cite{jung2016less}  presented a method for domain transfer learning. They try to maintain the performance on the old task by freezing the final layer of the old tasks and discouraging the shared weight parameters in the feature extraction layers from changing. \cite{kirkpatrick2017overcoming} proposed a technique to remember old tasks by constraining the important weight parameters to stay close to their old values while looking for a solution to a new task in the neighborhood of the old one. One limitation of the approach is that there will be conflicting constraints for the weight parameters which are shared between old and new tasks.  \cite{li2016learning} presented a method that uses a technique called knowledge distillation~\cite{hinton2015distilling} to maintain the performance of the new classifier on the old tasks.  \cite{Shmelkov-et-al-ICCV-2017} also uses knowledge distillation, but they apply it to the object detection instead of classification. Our method also belongs to this category in that we do not require the old data. The key difference is that we use a GANs \cite{goodfellow2014generative,ArjovskyCB17,RadfordMC15} to represent the old data. Even though the data generated by GANs are not as good as the real data, our experiments show that the data generated by GANs significantly improve the incremental learning performance compared to those techniques without using any old data at all.

The methods in the second category require part or all of the old data. \cite{rebuffi2016icarl} proposed a method to select a small number of exemplars from each old class, and the selected exemplars are stored for the incremental learning when adding new classes. \cite{lopez2017gradient}  proposed a continuous learning framework where the training samples for different tasks are input into the model one by one during training. \cite{xiao2014error}  proposed a training method that grows a network hierarchically as new training data are added. The key difference between our method and these techniques is that we do not keep any real data of the old classes. In real world applications, due to the privacy and legal concerns, real data may not be allowed to be stored for a long period of time.  Since GANs generate data purely from  a neural network without copying any real image patches, we expect GANs to have much less or even no privacy concerns. The situation is similar to the speech synthesis where the speech data generated by model-based speech synthesis methods has no privacy or legal concerns while the speech data generated by exemplar-based speech synthesis methods, which copies exemplar speech segments, are considered as having the same privacy concerns as the original real data.

\section{Incremental Classifier Learning \label{secincrementallearning}}

We define the incremental classification as follows: given a pre-trained classifier $\hat{f}^n$ on $n$ classes and a new labeled dataset for additional $m$ classes, how to train a new model $f^{n+m}$ to perform classification on $n+m$ classes? Let us denote the new dataset as $X^m=\{(x_i, y_i), 1 \leq i \leq M, y_i \in [n+1,..,n+m]\}$, where $M$ is the size of the dataset, $x_i$ and $y_i$ are the image and the label, respectively. 

Although strict incremental learning does not allow using the old $n$ class data, which is used to train the old model $\hat{f}^n$, iCaRL \cite{rebuffi2016icarl} shows that keeping a small amount of old data significantly improves the performance. Let us denote the selected old data as $\hat{X}^n=\{(\hat{x}_j, \hat{y}_j), 1 \leq j \leq N_s, \hat{y}_j \in [1,..,n]\}$, where $N_s$ is the number of selected old images.

\subsection{Design of Loss Function}
Incremental learning has two major challenges - (a) maintaining the classification performance on the old $n$ classes, and (b) balancing the old $n$ classes and the new $m$ classes. 

LwF \cite{li2016learning} used distilling with the new data to handle the first challenge and relied on the weight decay to address the balance. However, one limitation of this solution is that the distilling with the new data may not guarantee that the new classifier has similar predictions as the old classifier on the old data. In addition, finding the optimum weight decay is difficult.These two issues can be significantly improved by selecting a few real exemplars from the old data. We will provide a solution later. 

iCaRL \cite{rebuffi2016icarl} solved the first challenge by carefully selecting a few exemplars from the old data, and handling the unbalance issue by using binary entropy loss for each class. However, the loss function is not as effective as the cross entropy to handle the relationship between classes.

We propose a new loss function by leveraging the strength of both LwF \cite{li2016learning} and iCaRL \cite{rebuffi2016icarl}. We borrow the exemplar idea from iCaRL to keep a few exemplars and use them on both the distilling and cross entropy losses. The introduction of the old data improves the distilling by enabling the similar prediction between the old and new classifiers on the old $n$ classes. Although exemplars also help balance the old $n$ classes and new $m$ classes as the cross entropy is computed across all $n+m$ classes, it is still difficult to find the balance point since the new data variations are significantly larger. 
We present a simple representation of the unbalance - a scalar on the prediction of the new $m$ classes. 
%However, we observe that the unbalance can be 
%the unbalance between the old and new classes is consistent on validation set and test set even with a simple representation of the unbalance - a scalar on the prediction of new $m$ classes. 
In addition, we can simply estimate the unbalance on the validation set and apply it to the new classifier. Experimental results show that our approach not only outperforms the LwF and iCaRL, but also is not sensitive to the exemplar selection. Furthermore, our new loss function and unbalance estimation also works well on GANs generated data. We illustrate the difference between our model, LwF and iCaRL in Figure \ref{fig:compareFramework}.

\subsection{Loss Function}
\begin{figure}
	\begin{center}
		\includegraphics[width=1.1\linewidth]{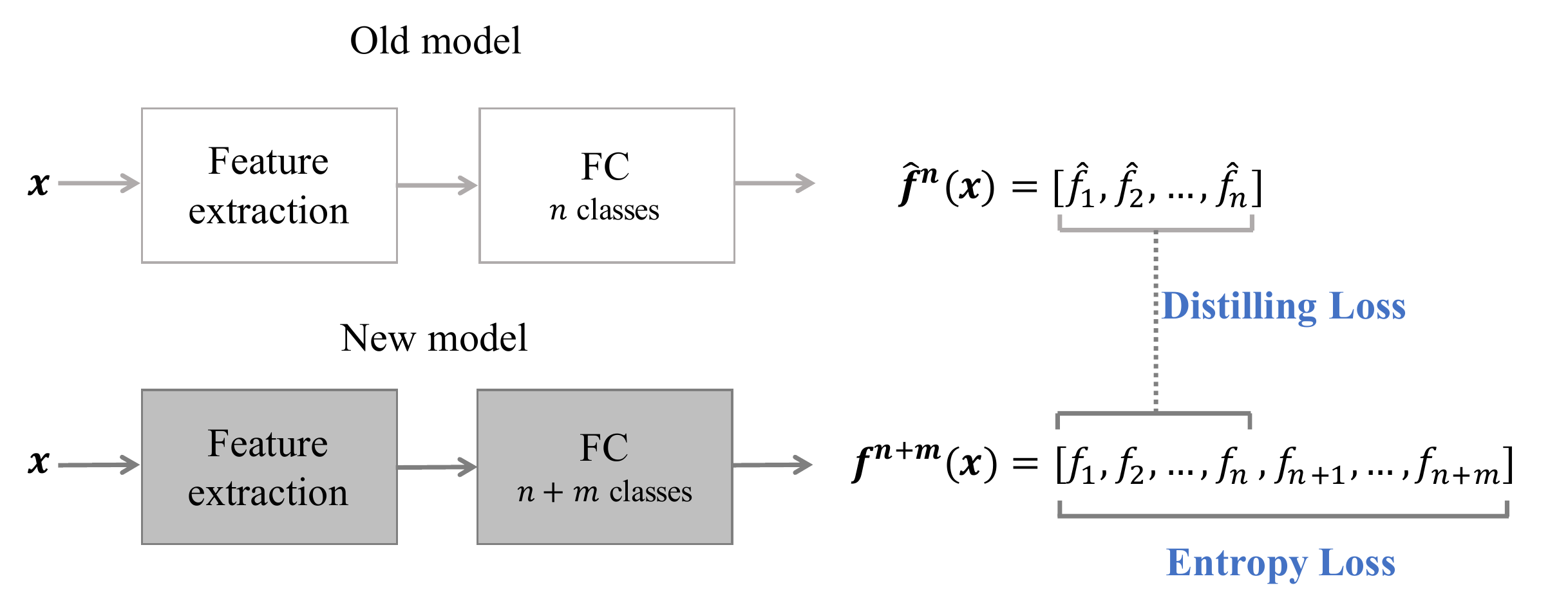}
	\end{center}
	%	\vspace{-4mm}
	\caption{Overview of our incremental learning framework. The loss function contains two terms, the distilling loss and the cross-entropy loss. }
	\label{fig:overview}
	%	\vspace{-4mm}
\end{figure}

In the section, we show the details of the proposed loss function and unbalance estimation. Let us denote the output of the old and new classifiers as $\hat{f}^n(x)=[\hat{f}_1(x),...,\hat{f}_n(x)]$ and $f^{n+m}(x)=[f_1(x),...,f_n(x), f_{n+1}(x),...,f_{n+m}(x)]$ respectively. This is illustrated in Figure \ref{fig:overview}.

The distilling loss is formulated as follows:
\begin{align}
L_d = \sum_{x\in \hat{X}^n  \cup X^m} \sum_{k=1}^n- \frac{\hat{f}_k(x)}{T}  \log (\frac{f_k(x)}{T}),
\end{align}
where $T$ is the temperature scaler and is set to 2. 

The cross-entropy loss is written as:
\begin{align}
L_c = \sum_{(x,y) \in \hat{X}^n \cup X^m} \sum_{k=1}^{n+m}-  \delta_{y=k} \log [f_k(x)].
\end{align}

The two losses are combined linearly as follows:
\begin{align}
L = \lambda L_d + (1-\lambda)L_c, \label{eqloss}
\end{align}
where the scaler $\lambda$ is used to balance the two terms. 

\subsection{Bias Removal}

Due to the unbalanced data between the old $n$ classes and the new $m$ classes (i.e. $N_s \ll M$), the new classifier $f^{n+m}$ is biased towards the new $m$ classes. This is confirmed by the experimental results. 
We simply represent the bias as 
%We found that the bias can be easily estimated on the validation set. 
%The bias is represented as
a multiplier $\beta$ between 0 and 1 that is applied to the outputs of the new $m$ classes:
\begin{align}
f^{n+m}(x)=[f_1(x),...,f_n(x), \beta f_{n+1}(x),...,\beta f_{n+m}(x)]. \label{eqbeta}
\end{align}
%We find this simple representation is consistent across the validation set and the test set. Thus, 
We can use the validation set to estimate the bias and apply it to the classifier to remove the bias. 
Results show that the bias is very consistent  across the validation set and the test set and our new loss with bias estimation outperforms the-state-of-arts iCaRL \cite{rebuffi2016icarl} on CIFAR-100.

\section{Generating Exemplars Using GANs \label{secsampling}}
The approach of choosing exemplars has a major drawback - the gap between the distribution of the chosen exemplars and the distribution of the entire training data for the old $n$ classes. 
Since in theory GANs \cite{goodfellow2014generative} can learn the distribution of the real dataset, GANs could potentially be a better representation of the old data. Furthermore, we expect GANs to have less privacy concerns than the real data because GANs do not copy any real image patches. Therefore, we propose training a GANs to represent the old data, and using the samples generated by the GANs for incremental learning.

\subsection{Combining GANs and Classifier}
We use all the training data of the old $n$ classes $\hat{X}^n$ to train a generator $x_g=G(z)$, where $z$ is a random noise input. Then a label is assigned to the generated image $x_g$ using the classifier $\hat{f}^n$ as follows:
\begin{align}
y_g = \underset{1 \leq k \leq n}{\arg\max}\hat{f}_k(x_g).
\end{align}
We use the vanilla GANs rather than the conditional GANs \cite{CGAN-2014} since it is difficult for the conditional GANs when the number of classes $n$ is large  (e.g. 10k for face data) and each class has limited samples \cite{ACGAN2017}. We use the DCGANs \cite{RadfordMC15} architecture for the generator and use the earth-mover distance in WGANs \cite{ArjovskyCB17} as the loss function to encourage more data variations and coverage.

\subsection{Data Selection \label{secganseect}} 
Even though there has been exciting progress in GANs, there are still limitations in the existing techniques. GANs cannot guarantee that the generated image 
$x_g$ always looks like images from the old training dataset $\hat{X}^n$. It sometimes generates images which are mixtures of multiple classes or generates images which are totally different from any of the classes. 
Therefore, a selection process is necessary in order to remove unrelated samples. We simply select generated images $x_g$ such that its maximum likelihood over $n$ classes is above a pre-specified threshold $\theta$, i.e. 
$\underset{1 \leq k \leq n}{\max}{\hat{f}_k(x_g)} > \theta$. Once the generated images are selected, we use the same incremental learning approach in Section \ref{secincrementallearning} to train the new classifier $f^{n+m}$ by replacing the real exemplars with generated images and labels $\{x_g, y_g\}$.

\section{Experiments}
In this section, we first introduce the datasets and describe implementation details. We then show comparisons with the state-of-the-art methods, followed by the discussions on the model parameters. 
\subsection{Datasets  }
Experiments are conducted on three datasets:\\ 
\textbf{CIFAR-100} \cite{krizhevsky2009learning} contains 100 object classes. Each class has 500 training images and 100 testing images. Following the class-incremental benchmark protocol in iCaRL \cite{rebuffi2016icarl} on this dataset, 100 classes are arranged in a fixed random order and come in as $P$ parts. Each part is  with $C=100/P$ classes. A multi-class classifier is built with the first part that contains $C$ classes.  Then this classifier is adapted to recognize all 100 classes. We have performed experiments for $P$=2, 5, 10 and 20.  \\
\textbf{Flower-102 } \cite{nilsback2008automated} consists of 102 flower categories. The original split of this dataset has 1,020 images in the training set and 6,149 images in the test set. To use more training images for the CNN model, we take the larger set of 6,149 images as training and test on the smaller set of 1,020 images. Similar to CIFAR-100, 102 classes are split into two parts and are trained incrementally. Each part contains 51 classes. \\
\textbf{MS-Celeb-1M-Base} \cite{lowshotface} has 20,000 classes with a total of 1.2 million aligned face images, which is a smaller  yet nearly noise-free version of MS-Celeb-1M \cite{guo2016msceleb} dataset. For each person, 80\%  with up to 30 images are randomly selected as the training images and the rest are used for testing. The 20,000 classes are divided randomly and equally into two parts to be incrementally trained. 

\subsection{Implementation Details  }
%and Implementation Details
We used the TensorPack package\footnote{\url{https://github.com/ppwwyyxx/tensorpack}} based on TensorFlow \cite{abadi2016tensorflow}. The implementation details are listed as follows:\\
\textbf{CIFAR-100}: we follow the iCaRL setting which used a 32-layer ResNet \cite{he2016deep}. Each training step has 70 epochs. The learning rate starts from 0.1 initially and reduces to 0.01 and 0.001 after 49 and 63 epochs, respectively. The weight decay is set to 0.0002 and the batchsize is 128. \\
\textbf{Flower-102}: an 18-layer ResNet \cite{he2016deep} is fine-tuned from a model that is trained with IMAGENET ILSVRC  2012 \cite{russakovsky2015imagenet} dataset. 90 epochs are used for each training. The learning rate is set to 0.1 and reduces to 1/10 of the previous learning rate every 30 epochs.  The weight decay is 0.0001 and the batchsize is 256. \\
\textbf{MS-Celeb-1M-Base}: a 34-layer ResNet \cite{he2016deep} is utilized. The learning rate, weight decay, batchsize and training epochs are the same as Flower-102. 
The standard Top-1 accuracy is reported as the evaluation metric. 

Our codes will be released in github.

\subsection{Comparisons and Results}
We compare with other alternative methods, including:\\
\textbf{Finetuning} learns an ordinary multi-class network without taking any measures to prevent catastrophic forgetting. It learns a multi-class classifier for new incoming classes by fine tuning the previously learned multi-class classification network.\\
\textbf{Learning without Forgetting (LwF)} \cite{li2016learning} utilizes distillation loss to prevent catastrophic forgetting.  But it does not use any method to recover the old data like iCaRL or ours. \\
\textbf{iCaRL} \cite{rebuffi2016icarl} keeps an exemplar set to store a small part of old data and uses distillation loss.  

The results of our method include:\\
\textbf{Ours-Real}  keeps a few exemplars like what iCaRL does but is trained with our loss function defined in Equation (\ref{eqloss}). \\
\textbf{Ours-GANs} uses GANs generated data to replace the exemplars in Ours-Real.  
The GANs model is trained by using the data in the first part, and the data in the first part are not used anymore during the incremental learning.

We compare Ours-Real to iCaRL when using real exemplars, and compare Ours-GANs to LwF and Finetuning without using exemplars.

\subsubsection{Ours-Real vs iCaRL with Real Exemplars}

We compare Ours-Real to iCaRL \cite{rebuffi2016icarl}, which is the state-of-arts on incremental learning with real exemplars.  The results for P=2 (C=50) are  shown in Table \ref{tablecifaricarl}. We can see that Ours-Real obtains 2.07\% gain in accuracy over iCaRL during increment learning from 50 classes to 100 classes. This demonstrates the superiority of our method. 

On CIFAR-100 dataset, ours-Real and iCaRL use different implementations of the 18-layer ResNet \cite{he2016deep}. When training classifier on all 100 class training data, our network gets 69.09\% top-1 accuracy, which is slightly higher than iCaRL (68.6\% accuracy reported in \cite{rebuffi2016icarl}). But when training classifier on the first 50 classes, our network obtains 75.22\% accuracy that is worse than 76.40\% in iCaRL. Given these two results, we consider that the two implementations are on par. Thus, the 2.07\% gain in accuracy shows that Ours-Real outperforms iCaRL comes from our loss function and unbalance removal. 

We further compare  iCaRL with Ours-Real on Flower-102 and MS-Celeb-1M-Base. Since results on these two datasets are not shown in \cite{rebuffi2016icarl}, we  implement their method by ourselves. Results are shown in Table \ref{talcompare}. 
	
On Flower-102, as suggested by \cite{rebuffi2016icarl}, the learning rate is initialized to 2 and decreased to 0.4 and 0.08 after 30 and 60 epochs, respectively. The total training epochs are 90. The size of the exemplar set is 255, which is the same as Ours-Real. From the results, we can see that Ours-Real outperforms iCaRL by 1\%. 
	
On MS-Celeb-1M-Base, the training with the sigmoid cross entropy loss on the old data of 10K classes does not converge regardless of whether we train it from scratch or fine-tune from the model that is trained using Softmax loss. 
This is probably because the sigmoid cross entropy loss does not enforce a strong discriminative boundary between classes.
Thus we adapt a two-stage fine-tuning strategy to make the  binary cross entropy loss  work on MS-Celeb-1M-Base. During the first stage, we fix the feature extraction  and only fine-tune the classifier with sigmoid cross entropy loss. After this stage, the top-1 accuracy on the validation set of the first 10k classes is 92\%. For the second stage, we fine-tune the whole network with a smaller learning rate for 40 epochs (using larger learning rates at the second stage does not converge).
After the second stage, we can achieve 97\% top-1 accuracy, which is still worse than the 99.34\% of the model trained with softmax. For the incremental learning with new data (another 10K classes),  the convergence problem still exists. Thus, we also adapt the same two-stage training strategy to make it converge. From the results, we can see that Ours-Real outperforms iCaRL by a margin, which shows that our method is better than iCaRL in handling the incremental learning with a large number of classes. 

\begin{table}[!t]
	\begin{center}
		\begin{tabular}{|l|c|c|c|c|c|}
			\hline
			Method &   CIFAR-100 &  Flower-102 & MS-Celeb\\
			\hline\hline
			iCaRL  \cite{rebuffi2016icarl}&0.6132 & 0.9301 &0.9341 \\
			Ours-Real  &  \textbf{0.6339} & \textbf{0.9402} &\textbf{0.9903} \\ 
			\hline
		\end{tabular}
	\end{center}
	%	\vspace{-4mm}
	\caption{Comparison of iCaRL and Ours-Real. Ours-Real outperforms iCaRL in all three datasets. On MS-Celeb-1M-Base that has 20,000 classes, Ours-Real is significantly better than iCaRL, which shows  that our incremental learning method is more effective in handling a large number of classes.} \label{talcompare}\label{tablecifaricarl}
	%	\vspace{-4mm}
\end{table}

The experimental results on CIFAR-100 with $P=5 (C=20), 10 (C=10)$, and $20 (C=5)$ are shown in Table \ref{tableicarl20}, \ref{tableicarl10} , and \ref{tableicarl5}, respectively. 
Our method outperforms iCaRL \cite{rebuffi2016icarl} for all incremental batches by a margin. 
The softmax outperforms sigmoid at the beginning by a large margin. Although the margin becomes smaller after a few batches, our method still has better performance across all batches. 
%The bias removal is critical to address class imbalance. Without it, the accuracy drops quickly below iCaRL \cite{rebuffi2016icarl} after a couple of batches. 
%Our method also outperforms iCaRL \cite{rebuffi2016icarl} on all batches for the splits of 10 and 20 classes. 
The average gains of our method over iCaRL \cite{rebuffi2016icarl} are $2.36$\%, $2.03$\% and $2.97$\% for the splits of 20, 10 and 5 classes, respectively. These results illustrate that our method using cross entropy and bias removal is effective and robust. 

\begin{table}[!t]
	\begin{center}
		\scalebox{1}{
			\begin{tabular}{|l|c|c|c|c|c|}
				\hline
				&   20 &  40 & 60 &   80 &  100\\
				\hline\hline
				iCaRL  \cite{rebuffi2016icarl}& 81.33&	72.19&{65.21}&{59.43} & {54.38}\\
				Ours-Real  &  \textbf{81.55} &\textbf{74.45} & \textbf{67.82} & \textbf{61.86}&  \textbf{56.53} \\ 
				%Ours-Real + center loss  &  81.35 &  74.03 & 66.03 & &\\ 
				
				\hline
			\end{tabular}
		}
	\end{center}
	%	\vspace{-4mm}
	\caption{Results on CIFAR-100 with a batch of 20 classes.} \label{tableicarl20}
	%	\vspace{-4mm}
\end{table}
\begin{table}[!t]
	\begin{center}
		\begin{tabular}{|l|c|c|c|c|c|c|c|c|c|c|}
			\hline
			&   10 &  20 & 30 &   40 &  50\\
			\hline
			iCaRL  \cite{rebuffi2016icarl}& 86.17&	78.60 & 73.60 &67.06 & 63.99 \\
			Ours-Real  &  \textbf{90.09}& \textbf{80.05} & \textbf{74.87}& \textbf{68.70} & \textbf{65.06} 	\\ 
			\hline
			
			\hline\hline
			&   60 &  70 & 80 &   90 &  100\\
			\hline
			iCaRL  \cite{rebuffi2016icarl}& 59.58 & 56.80 & 53.76 & 51.22 & 49.23 \\
			Ours-Real   & \textbf{61.70} & \textbf{59.13} & \textbf{56.03} & \textbf{53.16} & \textbf{51.38}	\\ 
			\hline
			
		\end{tabular}
	\end{center}
	%	\vspace{-4mm}
	\caption{Results on CIFAR-100 with a batch of 10 classes.} \label{tableicarl10}
	%	\vspace{-4mm}
\end{table}

\begin{table}[!t]
	\begin{center}
		
		\scalebox{1}{
			\begin{tabular}{|l|c|c|c|c|c|c|c|c|c|c|}
				\hline
				&   5 &  10 & 15 &   20 &  25\\
				\hline
				iCaRL  \cite{rebuffi2016icarl}& 87.80 & 82.50 & 78.06 &74.21 & 70.96 \\
				Ours-Real  &  \textbf{95.80}& \textbf{90.30} & \textbf{81.07}& \textbf{75.95} & \textbf{73.68} 	\\ 
				\hline
				
				\hline\hline
				&   30 &  35 & 40 &   45 &  50\\
				\hline
				iCaRL  \cite{rebuffi2016icarl}& 66.68&	64.16 & 62.18 &61.14 & 59.75 \\
				Ours-Real  &  \textbf{71.80}& \textbf{68.12} & \textbf{65.45}& \textbf{62.89} & \textbf{61.90} 	\\ 
				\hline
				
				\hline\hline
				&   55 &  60 & 65 &   70 &  75\\
				\hline
				iCaRL  \cite{rebuffi2016icarl}& 58.09&	56.58 & 54.73 &52.62 & 51.62 \\
				Ours-Real  &  \textbf{59.37  }& \textbf{57. 27} & \textbf{56.30}& \textbf{55.92} & \textbf{ 54.04} 	\\ 
				\hline
				
				\hline\hline
				&   80 &  85 & 90 &   95 &  100\\
				\hline
				iCaRL  \cite{rebuffi2016icarl}& 50.28 & 48.75 & 47.00 & 45.77 & 44.36 \\
				Ours-Real   & \textbf{52.63} & \textbf{49.75} & \textbf{49.68} & \textbf{48.16} & \textbf{46.91}	\\ 
				\hline
				
			\end{tabular}
		}
	\end{center}
	%	\vspace{-4mm}
	\caption{Results on CIFAR-100 with a batch of 5 classes.} \label{tableicarl5}
	%	\vspace{-4mm}
\end{table}

\begin{figure*}[!t]
	\begin{center}
		\includegraphics[width=1\linewidth]{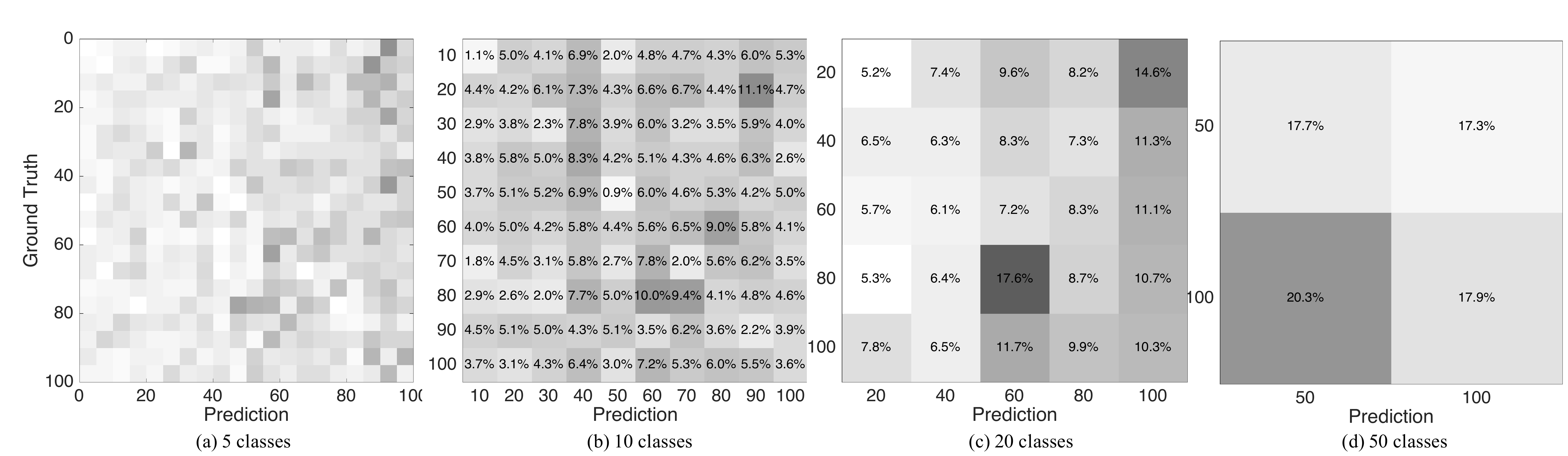}
	\end{center}
	\caption{Confusition matrix across batches, (a) split of 5 classes, (b) split of 10 classes, (c) split of 20 classes and (d) split of 50 classes.}
	\label{fig:long}
	\label{fig:confusion}
\end{figure*}

We  compare the confusion within and across  batches. Figure \ref{fig:confusion} shows the batch confusion on the split of 5, 10, 20 and 50 classes. The confusion is balanced inside and across batches overall.  
	
Another experiment is designed to compare our model with iCaRL \cite{rebuffi2016icarl} in the extreme case, i.e. using all old 50 classes data as exemplars. As shown in Table \ref{tableicarlalldata}, iCaRL and our model have similar performance in training a classifier in batch mode to recognize the 100 classes. 
In the incremental learning setting, our method can achieve the same performance compared with the batch training since the loss function for batch training is a special case of our method when $\lambda$ is set to 0. The incremental learning result of our method is actually slightly better due to the fact that the model of 50 classes is utilized as the initialization. Results of iCaRL cannot reach the upper bound of the batch training of its own. This is because the loss function in iCaRL during incremental learning is different to the batch training classifier. These results illustrate that our loss function (cross-entorpy loss with Softmax) is more efficient than the binary cross-entropy loss with a sigmoid activation function in iCaRL.

%\begin{table}
%	\begin{center}
%		\begin{tabular}{|l|c|c|c|c|c|}
%			\hline
%			Method &   CIFAR-100\\
%			\hline\hline
%			iCaRL  \cite{rebuffi2016icarl}&0.6132 \\
%			Ours-Real  &  \textbf{0.6339} \\ 
%			\hline
%		\end{tabular}
%	\end{center}
%	%	\vspace{-4mm}
%	\caption{Top-1 accuracy on CIFAR-100 with exemplars-based methods. An incremental step of 50 classes is utilized. 
%		Ours-Real outperforms iCaRL by 2.07\% accuracy.  
%	} \label{tablecifaricarl}
%	%	\vspace{-4mm}
%\end{table}

\begin{table}
	\begin{center}
		\begin{tabular}{|l|c|c|c|c|c|}
			\hline
			&Method & Top-1 accuracy \\
			\hline
			\multirow{ 2}{*}{Batch training} &iCaRL \cite{rebuffi2016icarl}&  0.6860\\
			&Ours-Real&  0.6909\\
			\hline
			\multirow{ 2}{*}{Incremental learning}&iCaRL \cite{rebuffi2016icarl}&  0.6687 \\
			&Ours-Real& 0.7004 \\
			\hline
		\end{tabular}
	\end{center}	
	%	\vspace{-4mm}
	\caption{Comparison between our method and iCaRL with all data available on CIFAR-100. Two methods achieve similar performance when using all the 100 classes at the batch training. Ours-Real outperforms iCaRL in the incremental learning setting.} \label{tableicarlalldata}
	%	\vspace{-4mm}
\end{table}

\begin{table}
	\begin{center}
		\begin{tabular}{|l|c|c|c|c|c|}
			\hline
			Method &     CIFAR-100&Flower-102 & MS-Celeb \\
			\hline\hline
			Finetuning &    0.3950 & 0.4912 & 0.4901 \\ 
			LwF  \cite{li2016learning}  &0.5250 & 0.7461&0.9848 \\ 
			Ours-GANs   & \textbf{0.5490} & \textbf{0.9000} & \textbf{0.9891}\\
			\hline\hline
			Ours-Real  & {0.6339}& {0.9402} &{0.9903}\\ 
			\hline
		\end{tabular}
	\end{center}
	%	\vspace{-4mm}
	\caption{Top 1 accuracy on CIFAR-100, Flower-102 and MS-Celeb datasets without using exemplars. Ours-GANs outperform LwF in all datasets.  
		Training with all data achieves top-1 accuracy 0.6909 on CIFAR-100, 0.9750 on Flower-102 and 0.9933 on MS-Celeb-1M-Base. 
		Results of Ours-Real are also given in the last row for reference. } \label{tableflowerface}
	%	\vspace{-4mm}
\end{table}

\begin{figure*}
	\begin{center}
		\includegraphics[width=\linewidth]{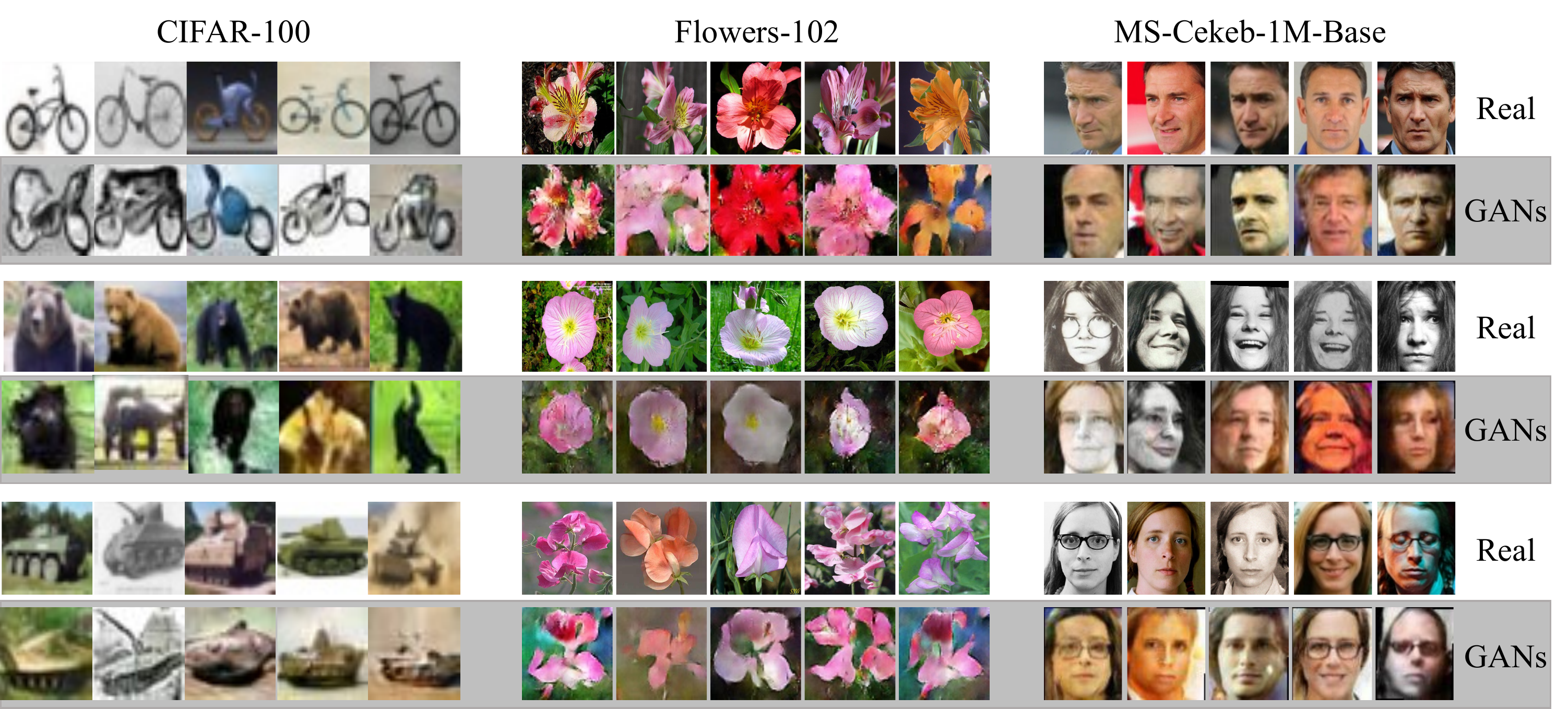}	
	\end{center}
	%	\vspace{-4mm}
	\caption{Illustration of some selected GANs images and real images of the corresponding class. We select three classes in each dataset and manually select the similar pair of images between the sampled GANs data and real images. Five pairs of every class are presented. Although the quality of GANs images are much lower than real images, some shapes and colors look similar between real images and GANs images.  }
	\label{figgandata}
	%	\vspace{-4mm}
\end{figure*}

\subsubsection{Ours-GANs vs LwF without Real Exemplars}

We compare Ours-GANs to LwF \cite{li2016learning}, which is the state-of-the-arts on incremental learning without any real exemplars. As shown in Table \ref{tableflowerface}, Ours-GANs outperforms LwF \cite{li2016learning} on all three datasets. 
Especially on Flower-102, Ours-GANs are 25.39\% better than LwF, which is a huge margin.  This shows the effectiveness our method and demonstrate that using distilling with the new data alone can not guarantee that the new classifier has similar predictions as the old classifier on the old data. . 

Compared with using the real images, the performance of using GANs drops from 63.87\% to 54.9\% on CIFAR-100. This is due to the challenge on training GANs on CIFAR-100 dataset. In Figure \ref{figgandata}, we can see that GANs generated images have poor quality. However, the generated images is still useful, since it provides better results (54.9\%) than LwF (52.5\%). We believe that as GANs continue to improve in the future, the gap between using real images and GANs will be reduced.

To visually verify that correlation between GANs generated data and real exemplars, we compare the real images and GANs generated images for three datasets in Figure \ref{figgandata}. For each dataset, we select 3 classes and 5 real images and GANs generated images per class. Clearly, the GANs generated images are correlated with the real images, which validates the idea of using GANs to represent the old training data. However, we also observe a clear difference between real images and GANs generated images, which explains the performance drop from using real exemplars to using GANs.

\subsection{Model Analysis}
In this subsection, we analyze the sensitivity of our loss function to (a) exemplar selection, (b) the scalar to balance cross entropy loss and distill loss $\lambda$ in Equation (\ref{eqloss}), and (c) the bias scalar $\beta$ in Equation (\ref{eqbeta}). The analysis is performed on CIFAR-100 dataset with an incremental from 50 classes to 100 classes. We also discuss how to sample the GANs generated data.

\subsubsection{Sensitivity to Exemplar-Selection}

We  evaluate our method with two different strategies to select real exemplars: (a) the exemplar management strategy proposed by iCaRL \cite{rebuffi2016icarl}, and (b) random selection. We keep the size of the exemplar set to 2,000. Random selection is run 5 times and an average result is reported. Results are shown in Table \ref{tableexemplar}. Compared with iCaRL exemplar management, randomly selected samples can achieve similar performance, which demonstrates that our loss function is  not sensitive to the exemplar-selection.

\begin{table}
	\begin{center}
		\begin{tabular}{|l|c|c|c|c|c|}
			\hline
			&Top-1 accuracy \\
			\hline
			iCaRL Exemplars \cite{rebuffi2016icarl}&  0.6339\\
			\hline
			Random Selection &  0.6361 \\
			\hline
		\end{tabular}
	\end{center}
	%	\vspace{-4mm}
	\caption{Comparison between iCaRL exemplar management strategy and random selection.}
	\label{tableexemplar}
	%	 \vspace{-4mm}
\end{table}

\subsubsection{Loss Scalar $\lambda$ }
The parameter $\lambda$ in Equation (\ref{eqloss}) controls the balance between the distillation loss and the cross-entropy loss. The value is set by grid search on a validation set.
Note that our validation data is split from the training set, not the test set.  

We first investigate the $\lambda$ with real data.  In Table \ref{tablelambdareal}, we can observe that the results are stable when varying $\lambda$ from 0.3 to 0.7. Thus we set $\lambda$ to 0.5. This setting is used for all the experiments with real data.  
The results show that the balance between the distilling loss and the cross entropy loss is important, since it provides the optimal trade-off between the old classifier and new data.
Note that when $\lambda$ is equal to 0.0, our loss function is equivalent to the cross-entropy loss. Using cross-entropy loss alone struggles with the data unbalance problem as it does not leverage the good performance of the old classifier. Using the distilling loss alone ($\lambda=$1.0), is not a good choice either since the labels of neither old nor new data are used.

Estimating $\lambda$ when using GANs generated data is difficult due to the apparent differences between the GANs generated images and the real images.
Intuitively, GANs data require a larger $\lambda$ compared with real images since outputs from the old classifier is more reliable than the pseudo labels. We empirically set $\lambda$ to 0.9 in all the experiments with GANs data.

\begin{table}
	\begin{center}
		\begin{tabular}{|l|c|c|c|c|c|}
			\hline
			$\lambda$ &  Validation & Test \\
			\hline
			1.0 & 0.482&   0.3703\\
			0.9 & 0.536&    0.5055 \\
			0.7 & 0.560 &  0.5576\\			
			0.5 & \textbf{0.562} & 0.566\\		
			0.3 & 0.558 & \textbf{0.5714} \\
			0.1 & 0.552 &  0.5546\\
			0.0 & 0.532 & 0.5352\\
			\hline
		\end{tabular}
	\end{center}
	%	\vspace{-4mm}
	\caption{Effect of $\lambda$ on the incremental learning performance when the exemplars are selected from the real old data set. } \label{tablelambdareal}
	%	\vspace{-4mm}
\end{table}

\begin{table}
	\begin{center}
		\begin{tabular}{|c|c|c||c|c|c|}
			\hline
			Old data & \multicolumn{2}{|c||}{Exemplars} & \multicolumn{2}{|c|}{GANs}\\
			\hline
			$\beta$ &  Validation & Test &  Validation & Test \\
			\hline
			0.0 & 0.4240 &   0.3594 &  0.4970 &   0.3740\\
			0.1 & 0.4240&    0.3594 & 0.5000&    0.3742\\
			0.2 & 0.4260&    0.3611 & 0.5080&    0.3848\\
			0.3 & 0.4420 &  0.3836& 0.5340 &  0.4130\\
			0.4 & 0.5040 &  0.4483& 0.5960 &  0.4574\\
			0.5 & 0.5680 & 0.5286& 0.6510 & 0.5042\\
			0.6 & 0.6400  & 0.6000& 0.7060  & 0.5472\\
			0.7 & \textbf{0.6720} & \textbf{0.6339}& 0.7400 & \textbf{0.5628}\\
			0.8 & 0.6580  & 0.6319 & \textbf{0.7570}  & 0.549\\
			0.9 & 0.6240 & 0.6064& 0.7490  & 0.5166\\
			1.0 & 0.5740 & 0.5667& 0.7140 & 0.4720\\
			\hline
		\end{tabular}
	\end{center}
	%	\vspace{-4mm}
	\caption{ 
		Incremental learning performance over $\beta$ on exemplars of real images and GANs generated data. The corresponding accuracy-$\beta$ curve is shown in Figure \ref{figexemplargancurve}. 
	} \label{tablebetarealReal}
	%	\vspace{-4mm}
\end{table}

\begin{figure}
	\begin{center}
		\includegraphics[width=0.9\linewidth]{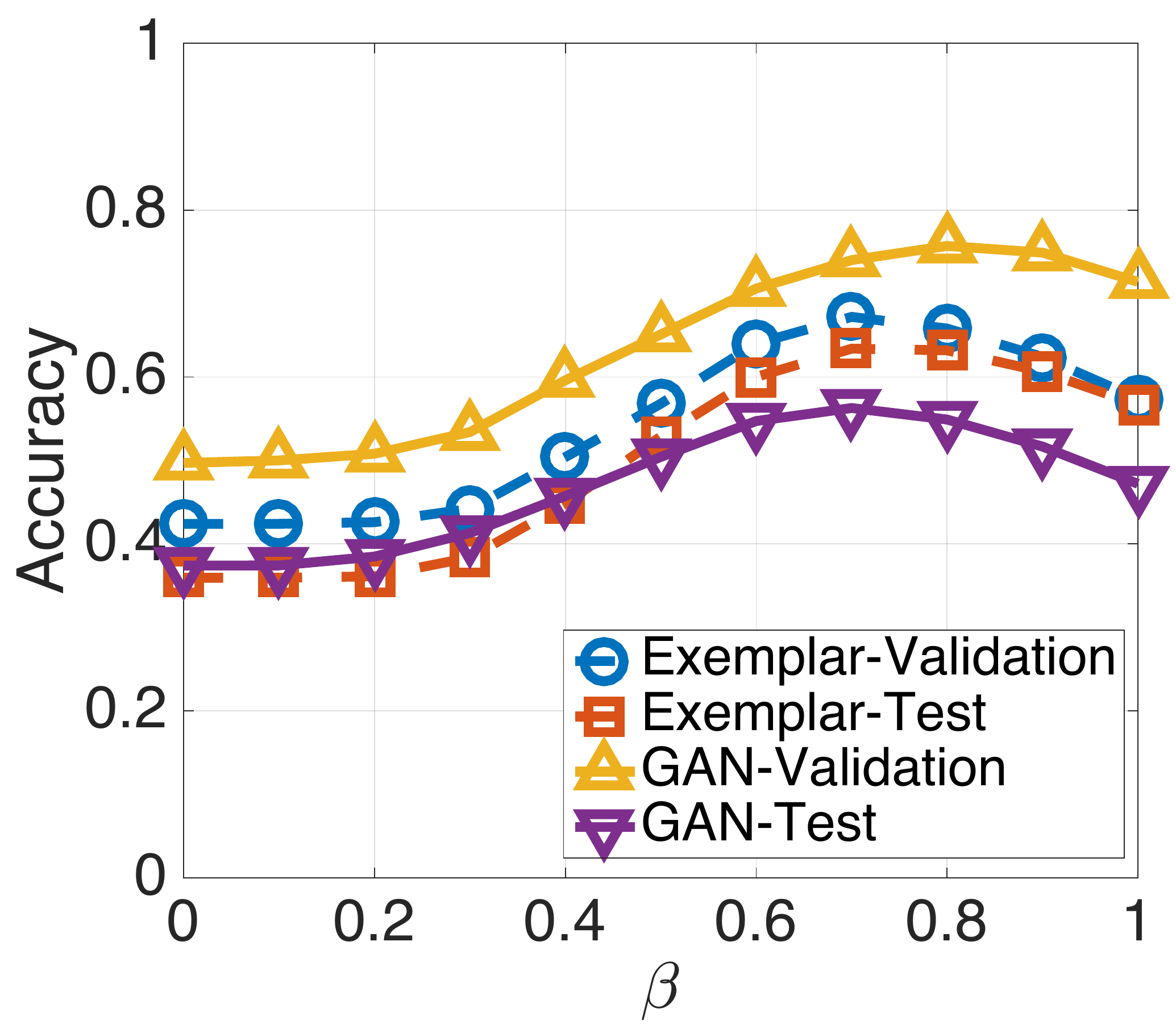}
		%		\vspace{-10mm}
	\end{center}
	%	\vspace{-4mm}
	\caption{ Incremental learning accuracy-$\beta$ curve. We can observe a correlation between the validation set and the test set with both real data and GANs generated data. }
	\label{figexemplargancurve}
	%	\vspace{-4mm}
\end{figure}

\begin{figure*}
	\begin{center}
		\includegraphics[width=\linewidth]{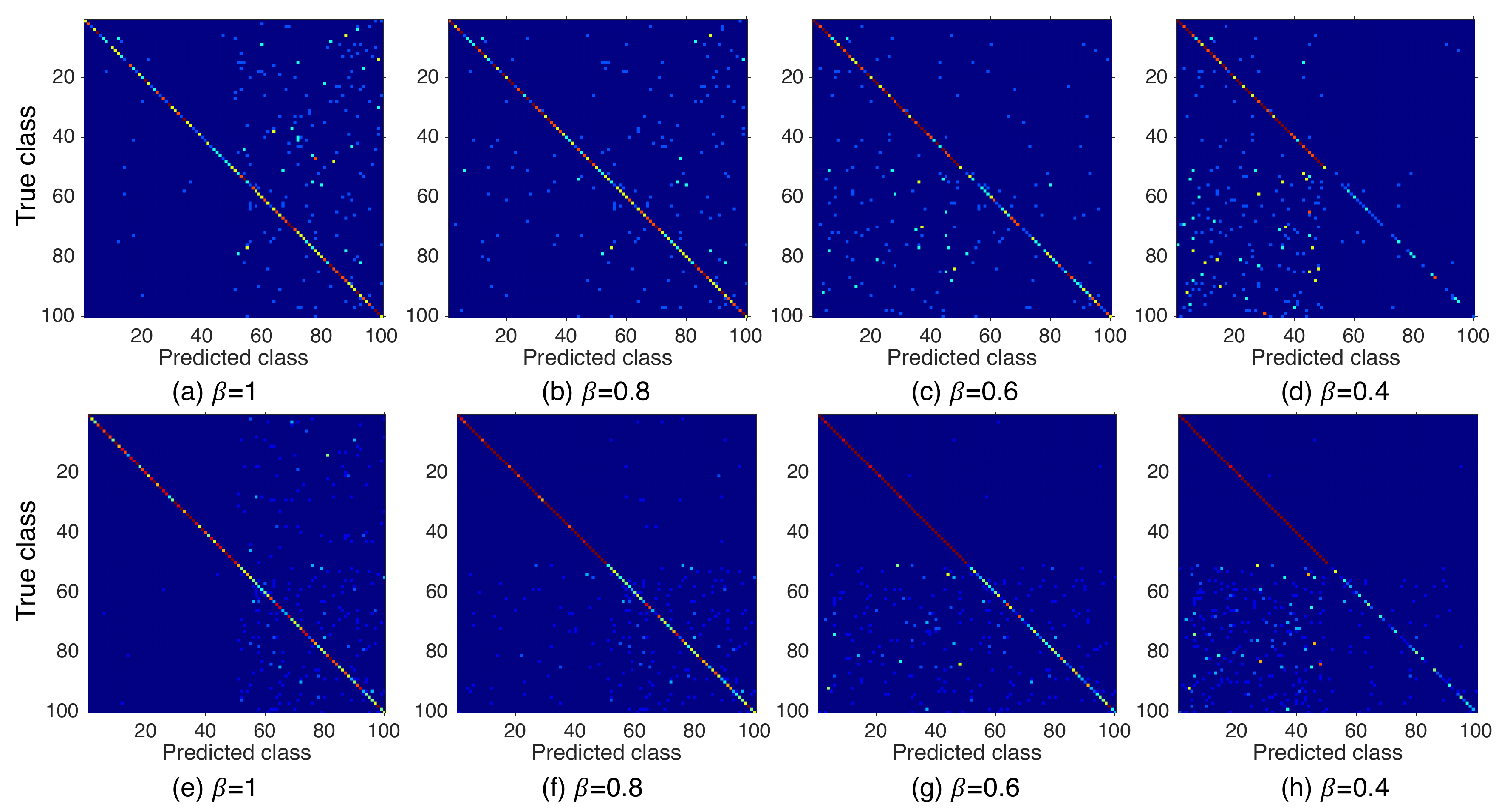}
		%		\vspace{-8mm}
	\end{center}
	\caption{Confusion matrix with different $\beta$ on the validation set with real data (a) (b) (c) (d) and GANs data (e) (f) (g) (h). We can observe that the bias moves gradually between the old 50 classes and the new 50 classes with different $\beta$. A proper $\beta$ can alleviate the bias in the classifier.}
	\label{figganval}
	%	\vspace{-4mm}
\end{figure*}

\subsubsection{Bias Scalar $\beta$}

The parameter $\beta$ in Equation (\ref{eqbeta}) is a multiplier applied to the output of new classes, which controls the balance between old classes and new classes. It can be estimated on the same validation set for estimating $\lambda$ in the last subsection.

We first analyze the effect of $\beta$ on  real image exemplars. 
We randomly select 5 samples per class from the training dataset for validation and keep the rest for training. 
For the first 50 classes, the 5 samples per class are selected from the 40 exemplars per class. For the newly arrived 50 classes, the 5 samples are selected from the training samples.
Results on the validation set is shown in Table \ref{tablebetarealReal} with $\beta$ varying from 0.0 to 1.0 with an increment of 0.1.  The accuracy-$\beta$ curve is shown in Figure \ref{figexemplargancurve}. We find that $\beta=0.7$ achieves the best with 9.8\% and 6.7\% gains on the validation set and the test set, respectively. This shows the effectiveness of our method to remove bias.

Compared to iCaRL, even though we use less data for training due to splitting into training set and validation,   
our bias-removed classifier outperforms iCaRL by 2.07\% in terms of accuracy,  shown in Table \ref{tableflowerface}.

We also analyze the selection of $\beta$ with GANs data.  
Similar to using the real exemplars, we select 10 samples from every class for validation. 
Results on the validation set and test set are shown in Table \ref{tablebetarealReal}. 
Using the best estimation of bias $\beta$ = 0.8 achieves 54.90\% top-1 accuracy, which is 7.7\% higher than the classifier without bias removal. We also observe that the validation and test sets are not perfectly aligned on the GANs generated data, i.e. a smaller bias $\beta$ = 0.7 achieves the best 56.28\% on the test dataset. This is because GANs generated data does not have the same distribution of the test dataset. But the bias estimation on the validation is reasonable with only 1.38\% drop in performance.

To further illustrate the bias problem in the incremental learning, we plot the confusion matrix with different $\beta$ on the validation set for both using exemplars and using GANs generated data in Figure \ref{figganval}.
We observe the same trend on both real exemplars and GANs generated data. When $\beta=1$, the bias is clearly on the new classes. As $\beta$ decreases, the bias gradually moves towards the old classes. We also observe a difference between using real exemplars and GANs generated data - the confusion between the old 50 classes is much smaller on the GANs generated data. This is due to a classic drawback of GANs: generated data have smaller variations than the real data.

\subsubsection{ GANs Data Sampling}
%We show how to set the parameters $\theta$ in section \ref{secsampling} and restrict the maximum number of selected  samples for each class in practice below. 
%GANs is trained on old data to generate samples. These samples are applied using the classifier learned from the same training data to assign a pseudo label along with a confidence score to each generated image. 
As described in section \ref{secsampling}, we utilize the threshold  $\theta$ to select good images. If the number of images with higher scores than $\theta$ is large, 
% be still a lot for some classes. 
%we specify a maximum number $K$ for the samples that are kept for each class. If a class has over $K$ samples, 
we select the top-$K$ images.  $\theta$ and $K$ are set empirically.  
On CIFAR-100, $\theta=0.95, K=50$. On Flower-102, $\theta=0.3, K=60$. On MS-Celeb-1M-Base, $\theta=0.3, K=10$. 

The $\theta$ on CIFAR-100 is very different to Flower-102 and MS-Celeb-1M-Base because the classifier is not as good as the other two. 

%Since the classifier is not powerful enough on CIFAR-100, e.g. about 75\% top-1 accuracy, we use 0.95 as the threshold $\theta$ and select samples with high confidences.  
%On Flower-102, and MS-Celeb-1M, $\theta$ is set to 0.3 given the powerful classifiers with over 95\% accuracy. For $K$, because the number of training images per class varies a lot in different datasets, e.g., 500 images in CIFAR-100, 30 images in MS-Celeb-1M-Base, we set $K$ to 50 on CIFAR-100 and 4 on MS-Celeb-1M-Base. On Flower-102, we set $K$ to 60 that is the same as the average number of training images. However, the intra-class variations of Flower-102 GANs data are not as diverse as real images in the same classes. Thus the unbalance problem still exists even with the same number of training images. 
%\balance
\section{Conclusions and Future Work}
In this paper, we address three issues in incremental classifier learning - (a) the inefficient loss function to integrate old classifier and new data, (b) the unbalance between old and new classes, (c) the scalable and privacy problems for selecting exemplars. 
Firstly, we propose a new loss function that combines the distillation loss and cross-entropy loss over all old and new classes. Secondly, we found that the unbalance between old and new classes can be represented as a multiplier on the prediction of new classes. This unbalance representation is very stable on validation set, test set and even data generated using GANs. Finally, we propose using GANs to generate the past data. Our method has excellent results outperforming the state of arts on three datasets CIFAR-100, FLower-102 and MS-Celeb-1M-Base by a large margin.
In the future, we plan to investigate how to improve the GANs generator to narrow the gap between GANs data and real images. 

%\clearpage
\balance

{\small
	\bibliographystyle{ieee}
	\bibliography{egbib}

\begin{thebibliography}{10}\itemsep=-1pt

\bibitem{abadi2016tensorflow}
M.~Abadi, A.~Agarwal, P.~Barham, E.~Brevdo, Z.~Chen, C.~Citro, G.~S. Corrado,
  A.~Davis, J.~Dean, M.~Devin, et~al.
\newblock Tensorflow: Large-scale machine learning on heterogeneous distributed
  systems.
\newblock {\em arXiv preprint arXiv:1603.04467}, 2016.

\bibitem{ArjovskyCB17}
M.~Arjovsky, S.~Chintala, and L.~Bottou.
\newblock Wasserstein generative adversarial networks.
\newblock In {\em Proceedings of the 34th International Conference on Machine
  Learning, {ICML} 2017, Sydney, NSW, Australia, 6-11 August 2017}, pages
  214--223, 2017.

\bibitem{cauwenberghs2001incremental}
G.~Cauwenberghs and T.~Poggio.
\newblock Incremental and decremental support vector machine learning.
\newblock In {\em Advances in neural information processing systems}, pages
  409--415, 2001.

\bibitem{goodfellow2014generative}
I.~Goodfellow, J.~Pouget-Abadie, M.~Mirza, B.~Xu, D.~Warde-Farley, S.~Ozair,
  A.~Courville, and Y.~Bengio.
\newblock Generative adversarial nets.
\newblock In {\em Advances in neural information processing systems}, pages
  2672--2680, 2014.

\bibitem{lowshotface}
Y.~Guo and L.~Zhang.
\newblock One-shot face recognition by promoting underrepresented classes.
\newblock {\em arXiv preprint arXiv:1707.05574}, 2017.

\bibitem{guo2016msceleb}
Y.~Guo, L.~Zhang, Y.~Hu, X.~He, and J.~Gao.
\newblock M{S}-{C}eleb-1{M}: A dataset and benchmark for large scale face
  recognition.
\newblock In {\em European Conference on Computer Vision}. Springer, 2016.

\bibitem{he2016deep}
K.~He, X.~Zhang, S.~Ren, and J.~Sun.
\newblock Deep residual learning for image recognition.
\newblock In {\em Proceedings of the IEEE conference on computer vision and
  pattern recognition}, pages 770--778, 2016.

\bibitem{hinton2015distilling}
G.~Hinton, O.~Vinyals, and J.~Dean.
\newblock Distilling the knowledge in a neural network.
\newblock In {\em NIPS Deep Learning and Representation Learning Workshop},
  2015.

\bibitem{jung2016less}
H.~Jung, J.~Ju, M.~Jung, and J.~Kim.
\newblock Less-forgetting learning in deep neural networks.
\newblock {\em arXiv preprint arXiv:1607.00122}, 2016.

\bibitem{kirkpatrick2017overcoming}
J.~Kirkpatrick, R.~Pascanu, N.~Rabinowitz, J.~Veness, G.~Desjardins, A.~A.
  Rusu, K.~Milan, J.~Quan, T.~Ramalho, A.~Grabska-Barwinska, et~al.
\newblock Overcoming catastrophic forgetting in neural networks.
\newblock {\em Proceedings of the National Academy of Sciences},
  114(13):3521--3526, 2017.

\bibitem{krizhevsky2009learning}
A.~Krizhevsky and G.~Hinton.
\newblock Learning multiple layers of features from tiny images.
\newblock 2009.

\bibitem{kuzborskij2013n}
I.~Kuzborskij, F.~Orabona, and B.~Caputo.
\newblock From n to n+ 1: Multiclass transfer incremental learning.
\newblock In {\em Proceedings of the IEEE Conference on Computer Vision and
  Pattern Recognition}, pages 3358--3365, 2013.

\bibitem{li2016learning}
Z.~Li and D.~Hoiem.
\newblock Learning without forgetting.
\newblock In {\em European Conference on Computer Vision}, pages 614--629.
  Springer, 2016.

\bibitem{lopez2017gradient}
D.~Lopez-Paz et~al.
\newblock Gradient episodic memory for continual learning.
\newblock In {\em Advances in Neural Information Processing Systems}, pages
  6470--6479, 2017.

\bibitem{McCloskey-Cohen-PLM-1989}
M.~McCloskey and N.~J.Cohen.
\newblock Catastrophic interference in connectionist networks: The sequential
  learning problem.
\newblock {\em Psychology of Learning and Motivation}, 24:109--165, 1989.

\bibitem{CGAN-2014}
S.~O. Mehdi~Mirza.
\newblock Conditional generative adversarial nets.
\newblock In {\em Deep Learning Workshop NIPS 2014}, 2014.

\bibitem{mensink2013distance}
T.~Mensink, J.~Verbeek, F.~Perronnin, and G.~Csurka.
\newblock Distance-based image classification: Generalizing to new classes at
  near-zero cost.
\newblock {\em IEEE transactions on pattern analysis and machine intelligence},
  35(11):2624--2637, 2013.

\bibitem{nilsback2008automated}
M.-E. Nilsback and A.~Zisserman.
\newblock Automated flower classification over a large number of classes.
\newblock In {\em Computer Vision, Graphics \& Image Processing, 2008.
  ICVGIP'08. Sixth Indian Conference on}, pages 722--729. IEEE, 2008.

\bibitem{ACGAN2017}
A.~Odena, C.~Olah, and J.~Shlens.
\newblock Conditional image synthesis with auxiliary classifier gans.
\newblock In {\em International Conference on Machine Learning}, pages
  2642--2651, 2017.

\bibitem{polikar2001learn++}
R.~Polikar, L.~Upda, S.~S. Upda, and V.~Honavar.
\newblock Learn++: An incremental learning algorithm for supervised neural
  networks.
\newblock {\em IEEE transactions on systems, man, and cybernetics, part C
  (applications and reviews)}, 31(4):497--508, 2001.

\bibitem{RadfordMC15}
A.~Radford, L.~Metz, and S.~Chintala.
\newblock Unsupervised representation learning with deep convolutional
  generative adversarial networks.
\newblock In {\em arXiv preprint arXiv:1511.06434 [cs.LG]}, 2015.

\bibitem{rebuffi2016icarl}
S.-A. Rebuffi, A.~Kolesnikov, G.~Sperl, and C.~H. Lampert.
\newblock icarl: Incremental classifier and representation learning.
\newblock In {\em The IEEE Conference on Computer Vision and Pattern
  Recognition (CVPR)}, July 2017.

\bibitem{russakovsky2015imagenet}
O.~Russakovsky, J.~Deng, H.~Su, J.~Krause, S.~Satheesh, S.~Ma, Z.~Huang,
  A.~Karpathy, A.~Khosla, M.~Bernstein, et~al.
\newblock Imagenet large scale visual recognition challenge.
\newblock {\em International Journal of Computer Vision}, 115(3):211--252,
  2015.

\bibitem{Shmelkov-et-al-ICCV-2017}
K.~Shmelkov, C.~Schmid, and K.~Alahari.
\newblock Incremental learning of object detectors without catastrophic
  forgetting.
\newblock In {\em Proceedings of the International Conference on Computer
  Vision}, 2017.

\bibitem{xiao2014error}
T.~Xiao, J.~Zhang, K.~Yang, Y.~Peng, and Z.~Zhang.
\newblock Error-driven incremental learning in deep convolutional neural
  network for large-scale image classification.
\newblock In {\em Proceedings of the 22nd ACM international conference on
  Multimedia}, pages 177--186. ACM, 2014.

\end{thebibliography}
}

\end{document}